\documentclass[10pt,twocolumn,letterpaper]{article}

\usepackage[pagenumbers]{cvpr} 

\usepackage{microtype}
\usepackage{graphicx}
\usepackage{amsmath}
\usepackage{amssymb}
\usepackage{booktabs}
\usepackage{dsfont}
\usepackage{wrapfig}
\usepackage{pifont}
\usepackage{threeparttable}
\usepackage{multirow}
\usepackage{makecell}
\usepackage[table]{xcolor}
\usepackage{xfrac}

\usepackage[pagebackref,breaklinks,colorlinks]{hyperref}

\usepackage{soul} 
\usepackage{xcolor}

\definecolor{pink}{rgb}{1,0.03,0.5}

\newcommand{\supp}[1]{\textcolor{black}{#1}}

\definecolor{darkgreen}{HTML}{4E9A3C}
\definecolor{darkred}{HTML}{BA3132}

\newcommand{\cmark}{\textcolor{darkgreen}{\ding{51}}}
\newcommand{\xmark}{\textcolor{darkred}{\ding{55}}}

\usepackage{enumitem}
\setlist[itemize]{noitemsep, topsep=0.1in}
\setlist[enumerate]{noitemsep, topsep=0.1in}

\usepackage[capitalize]{cleveref}
\crefname{section}{Sec.}{Secs.}
\Crefname{section}{Section}{Sections}
\Crefname{table}{Table}{Tables}
\crefname{table}{Tab.}{Tabs.}

\newcommand{\HDRGAN}{GlowGAN}

\DeclareMathOperator*{\argmin}{argmin}

\begin{document}

\title{\HDRGAN{}: Unsupervised Learning of HDR Images from \\ LDR Images in the Wild}

\author{Chao Wang$^1$, Ana Serrano$^2$, Xingang Pan$^1$, Bin Chen$^1$, Hans-Peter Seidel$^1$\\
 Christian Theobalt$^1$, Karol Myszkowski$^1$, Thomas Leimk\"uhler$^1$
 \vspace{0.3cm}
\\
$^1$ Max Planck Institute for Informatics, Germany,  $^2$ Universidad de Zaragoza, Spain\\
}

\twocolumn[{
  \renewcommand\twocolumn[1][]{#1}
  \maketitle
  \vspace{-10mm}
  \begin{center}
    \includegraphics[width=\textwidth]{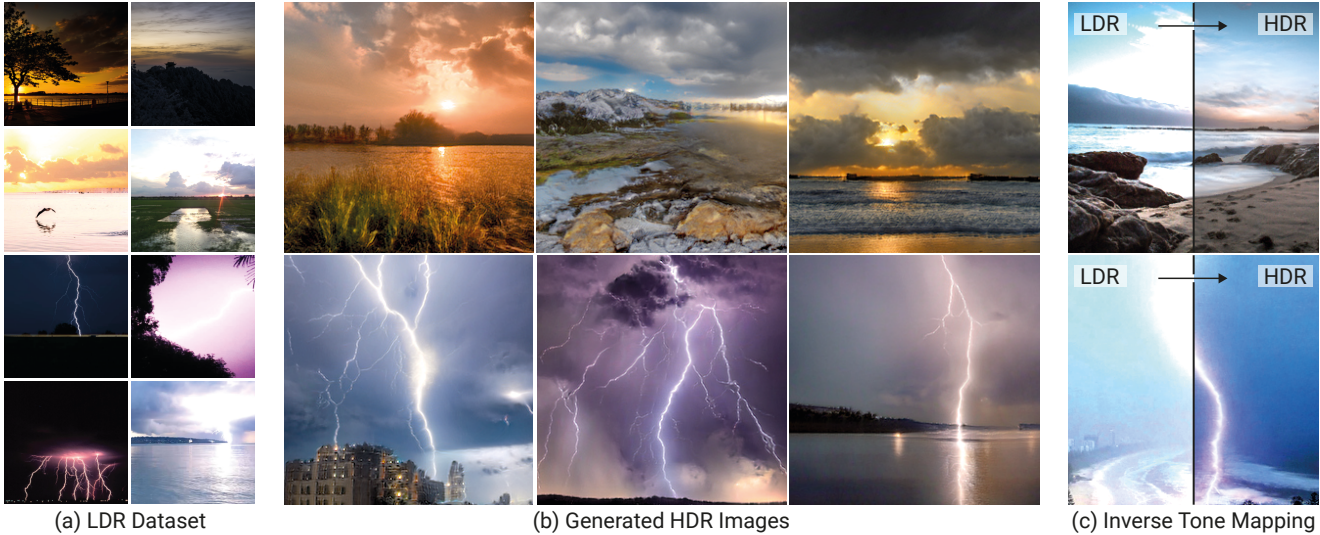}
  \end{center}
  \vspace{-5mm}
  \captionof{figure}{Typical image datasets have a low dynamic range (LDR), resulting in over- and underexposed pixels (\emph{a}). We devise a high-dynamic-range (HDR) generator trained only on in-the-wild LDR data. Our HDR samples (\emph{b}, tonemapped for display) exhibit details at all brightness levels. Our model can be used for inverse tone mapping to recover details in large-scale saturated regions (\emph{c}).}
  \label{fig:teaser}
  \vspace{4mm}
}]

\begin{abstract}
Most in-the-wild images are stored in Low Dynamic Range (LDR) form, serving as a partial observation of the High Dynamic Range (HDR) visual world.
Despite limited dynamic range, these LDR images are often captured with different exposures, implicitly containing information about the underlying HDR image distribution.
Inspired by this intuition, in this work we present, to the best of our knowledge, the first method for learning a generative model of HDR images from in-the-wild LDR image collections in a fully unsupervised manner.
The key idea is to train a generative adversarial network (GAN) to generate HDR images which, when projected to LDR under various exposures, are indistinguishable from real LDR images.
The projection from HDR to LDR is achieved via a camera model that captures the stochasticity in exposure and camera response function.
Experiments show that our method \HDRGAN{} can synthesize photorealistic HDR images in many challenging cases such as landscapes, lightning, or windows, where previous supervised generative models produce overexposed images.
We further demonstrate the new application of unsupervised inverse tone mapping (ITM) enabled by \HDRGAN{}. Our ITM method does not need HDR images or paired multi-exposure images for training, yet it reconstructs more plausible information for overexposed regions than state-of-the-art supervised learning models trained on such data. 
\end{abstract}

\section{Introduction}
\label{sec:intro}

High Dynamic Range (HDR) images \cite{reinhard2010high} are capable of capturing and displaying much richer appearance information than Low Dynamic Range (LDR) images, thus playing an important role in image representation and visualization.
The most popular method to acquire HDR images is multiple exposure blending, which requires capturing a set of LDR images of the same scene with different exposures~\cite{debevec1997recovering, Mitsunaga1999RadiometricSC,robertson2003estimation}. 
However, this is time and effort intensive and only suitable for static scenes. 
Due to this limitation, existing HDR image datasets only cover limited scene categories and have much fewer images than LDR datasets.
Thus, supervised learning methods ~\cite{eilertsen2017hdr, marnerides2018expandnet, endo2017deep, lee2018deep, liu2020single, lee2020learning, santos2020single, zhang2021deep, yu2021luminance, jo2021deep} that reconstruct an HDR image from an LDR image are constrained by the HDR datasets and cannot extend to cases where no HDR training data is available, e.g., lightning.

While HDR images are hard to collect, it is much easier to collect a large number of LDR images from the Internet. 
This motivates us to investigate a new unsupervised learning problem: \textit{Can we learn to reconstruct HDR images from in-the-wild LDR images?}
The LDR images do not need to depict the same scene, it is enough if they contain a roughly similar class of scenes (\eg, landscapes) with various exposures.
This weak multi-exposure assumption is often naturally satisfied for in-the-wild LDR datasets as images can come from different camera parameters or different adjustments of the auto-exposure mode.
This problem is challenging as only one exposure is available for each scene, therefore, a way to merge the multi-exposure information spread across different scenes is required.

In this work, we address this challenge via \textit{\HDRGAN{}}, which, to our knowledge, is the first method to learn an HDR generative model from in-the-wild LDR image collections in a fully unsupervised manner.
\HDRGAN{} uses adversarial training of an HDR generator with a discriminator that operates merely in LDR.
Specifically, the generator produces an HDR image, which is projected to LDR via a camera model and is then sent to the discriminator as a fake image for adversarial training.
The camera model consists of multiplying the HDR sample with an exposure value, clipping the dynamic range, and applying a camera response function (CRF).
Importantly, during training, we use a randomly sampled exposure from a prior Gaussian distribution when projecting HDR to LDR.
This requires the generated HDR images to be realistic under any possible exposure, thus only valid HDR samples would satisfy this ``multi-exposure constraint''.
Furthermore, we also model the stochasticity in the non-linear camera response by randomly sampling CRFs according to a well-established parametric distribution \cite{eilertsen2017hdr, grossberg2003space}. 
During the inference process, we can disable the camera model so that the generator produces HDR imagery directly.

We conduct extensive experiments on several datasets collected from the Internet, including landscapes, windows, lightning, fireplaces, and fireworks.
By training on these LDR images, \HDRGAN{} successfully learns to generate high-quality HDR images that capture rich appearance information from dark to very bright regions.
These details can be presented on HDR displays, or via suitable tone mapping to create appealing imagery.
In contrast, previous unconditional GANs tend to miss information in over- or under-exposed regions.

By modelling a distribution of HDR images, \HDRGAN{} paves the way for new applications such as unsupervised inverse tone mapping (ITM).
ITM aims to reconstruct an HDR image from a single-exposure LDR input, where a key challenge is to restore the flat-white overexposed regions ~\cite{eilertsen2017hdr, santos2020single}.
We can use a pre-trained \HDRGAN{} as a prior and apply GAN inversion to optimize latent code and exposure, making the model generate the corresponding HDR image for the input LDR image.
An exciting result is that our method can, without using any HDR imagery or paired multi-exposure data, reconstruct starkly more plausible information for large overexposed regions than other supervised learning approaches trained on such data.
Furthermore, the HDR samples generated by \HDRGAN{} can be used as versatile environment maps in rendering.
%
Our contributions are summarized as follows:
\begin{itemize}
    \item We are the first to present unsupervised learning of HDR images from in-the-wild LDR images.  
    This gets rid of the reliance on ground truth HDR images that are much harder to collect.
    \item To achieve this, we propose a novel \HDRGAN{}, which bridges HDR space and LDR space via a camera model.
    \HDRGAN{} can synthesize diverse high-quality images with a much higher dynamic range than vanilla GANs, opening up new avenues for getting cheap abundant HDR data.
	\item Using \HDRGAN{} as a prior, we design an unsupervised inverse tone mapping method (ITM), which reconstructs large overexposed regions significantly better than the state-of-the-art fully-supervised approaches.
\end{itemize}
The supplementary material is provided in \href{https://drive.google.com/drive/folders/1b1XDpBE-V_Xc-QLtXbEnxjb03ZloCbFb?usp=sharing}{this link}. 
Our code and pretrained models will be released.

\section{Related Work}
\label{sec:related_work}

\subsection{High Dynamic Range Imaging}

The real world has a vast dynamic range. Therefore, HDR imaging~\cite{reinhard2010high} is crucial for creating and manipulating immersive viewing experiences. While HDR capture is cumbersome and HDR displays are not yet commonplace in most environments~\cite{seetzen2004high}, the representation of visual information free from the limitations of typical LDR encodings has evolved significantly in recent years~\cite{wang2021deep}. Working with HDR content is crucial for rendering~\cite{debevec2002image} and has been shown to be beneficial for 3D scene reconstruction: An HDR representation can naturally handle multi-exposure~\cite{jun2022hdr,huang2022hdr} and raw data~\cite{mildenhall2022nerf}. Reconstructing HDR in conjunction with an explicit tone mapping module can compensate for poorly calibrated cameras~\cite{ruckert2022adop}.

A particular interest has evolved around the conversion between HDR and LDR content. Tone mapping, the transformation from HDR to LDR with as little information loss as possible, is a mature field with well-understood trade-offs~\cite{reinhard2010high, wang2022learning}. In contrast, inverse tone mapping (ITM)~\cite{banterle2006inverse,rempel2007ldr2hdr}, the recovery of HDR content from LDR imagery, remains a challenging inverse problem. It typically involves multiple steps, including linearization, dynamic range expansion, reconstruction of over- and underexposed regions, artifact reduction, and color correction~\cite{banterle2017advanced}. Among these, the reconstruction of saturated pixels is considered the most challenging~\cite{eilertsen2017hdr, santos2020single}, as it requires the hallucination of content~\cite{wang2007high}.

Early ITM works considered expansion curves using either global~\cite{landis2002production, masia2009evaluation, akyuz2007hdr, masia2017dynamic} or content-driven operators~\cite{meylan2006reproduction, banterle2006inverse, banterle2009psychophysical, banterle2008expanding, rempel2007ldr2hdr,rempel2007ldr2hdr}, without explicitly reconstructing saturated regions. More recent learning-based solutions can be categorized into two streams: A neural network either predicts the HDR image directly ~\cite{eilertsen2017hdr, marnerides2018expandnet, santos2020single, liu2020single, zhang2021deep, yu2021luminance,chen2022text2light}, or it predicts multiple LDR images with different exposures~\cite{endo2017deep,lee2018deep,lee2018re,lee2020learning,jo2021deep}, which are subsequently merged into an HDR image~\cite{debevec1997recovering, Mitsunaga1999RadiometricSC,robertson2003estimation}. ITM and exposure fusion can be combined with adversarial training~\cite{zhang2021deep, lee2018re,niu2021hdr}. Also, the extension to video ITM has been explored extensively, leveraging inter-frame consistency~\cite{kalantari2019deep, kim2019deep, kim2020jsi, banterle2022unsupervised, he2022global, he2022sdrtv}.

All learning-based techniques discussed above rely on supervision from paired LDR--HDR training data. This constitutes a fundamental problem: HDR image data is hard to obtain and therefore naturally scarce. Further, most HDR capture techniques require static content, which significantly restricts the applicability of learning-based methods to arbitrary scenes. In contrast, we are the first to train an HDR image generator \emph{unsupervised} from an LDR dataset. We believe this is an important step towards solving two main problems in HDR imaging: First, our generator can synthesize an abundance of HDR samples, alleviating the scarcity of HDR content. Second, our system allows to perform ITM with a significant improvement in the reconstruction of overexposed regions.

\subsection{Lossy Generative Adversarial Networks}

Generative Adversarial Networks (GANs)~\cite{goodfellow2014generative} are very successful in modelling distributions of images with high visual fidelity. The StyleGAN family~\cite{karras2019style, karras2020analyzing, karras2020training, karras2021alias} marks the current state of the art, scaling to high resolutions, while the recent extension StyleGAN-XL~\cite{sauer2022stylegan} allows for unprecedented diversity in the generated content. To this date, (unconditional) GANs operate in LDR, since high-quality HDR data at the scale required for successful training is difficult to obtain. We propose a simple modification to the GAN training pipeline, which allows to train an HDR generator from readily available LDR data only.

AmbientGAN~\cite{bora2018ambientgan} has demonstrated that it is feasible to train a generative model from lossy measurements, i.e., a GAN can be trained from degraded samples, as long as the stochastic properties of the degradation are known. This concept has been used to learn a generator for clean images from noisy data~\cite{kaneko2020noise}, or for all-in-focus images from data that contains shallow depth of field~\cite{kaneko2021unsupervised}. Most prominently, the idea has been applied to learn 3D generators from 2D images by explicitly modeling the projection from 3D to 2D using a distribution of extrinsic camera parameters~\cite{henzler2019escaping,chan2021pi,liao2020towards,nguyen2019hologan,schwarz2020graf,szabo2019unsupervised}. We follow the paradigm of injecting a degradation model into the GAN training pipeline, by devising a novel model of the distribution of processing steps in a digital camera, converting HDR radiance into LDR pixel intensities.

\section{Method}

\begin{figure*}
	\centering
 \includegraphics[width=0.9\textwidth]{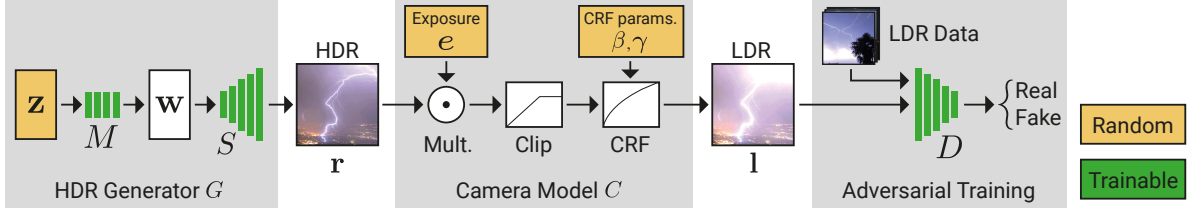}
 \vspace{-0.1cm}
	\caption{Overview of \HDRGAN{}. The generator generates an HDR image $\mathbf{r}$ from a random noise $\mathbf{z}$. Then a camera model $C$ projects $\mathbf{r}$ to an LDR image $\mathbf{l}$ with a random exposure and CRF. The model is trained merely on in-the-wild LDR images in an adversarial manner.} 
 \vspace{-0.3cm}
	\label{Fig.structure} 
\end{figure*}

We train a GAN~\cite{goodfellow2014generative} to capture the distribution of HDR images in a domain (e.g., landscapes) by combining it with a stochastic camera model that transforms the generated HDR images into their LDR counterparts. The discriminator is only fed LDR images, which allows the system to be trained on an easily accessible in-the-wild LDR image dataset. Our camera model can be inserted into any GAN model to yield HDR outputs as long as the LDR training dataset consists of images exhibiting different exposures across samples. We consider this a mild assumption, which in particular in-the-wild photo datasets easily satisfy. See Fig.~\ref{Fig.structure} for an overview of our system.
In the following, we describe our pipeline in detail (Sec.~\ref{sec-pipeline}), before turning to our main application of unsupervised inverse tone mapping (Sec.~\ref{sec-itm}).

\subsection{GlowGAN}
\label{sec-pipeline}

We seek to capture the unknown true distribution of HDR images $p_\textrm{HDR}$ from samples of the distribution of LDR images $p_\textrm{LDR}$. Similar to the standard GAN setup, we achieve this by training a generator $G$ which turns a random latent vector $\mathbf{z} \in \mathds{R}^k$ into an HDR sample $\mathbf{r} \in \mathds{R}_{\geq 0}^{H \times W \times 3}$, an RGB image with $H \times W$ pixels and no upper restriction on the value range.
To train $G$, we inject a camera model 
$C \in \mathds{R}_{\geq 0}^{H \times W \times 3} \rightarrow [0, 1]^{H \times W \times 3}$ 
into the adversarial training pipeline, turning the unbounded HDR image into an LDR image with values between 0 and 1.
$C$ captures the distribution $p_\textrm{cam}$ of pixel-wise image transformations typically applied in a digital camera to convert incoming radiance values to final pixel intensities, including varying exposures, clipping, and varying non-linearities arising from the camera response function (CRF). 
The result of this process is an LDR image 
$\mathbf{l} = C(\mathbf{r}) = C(G(\mathbf{z}))$.
The discriminator $D$ is tasked with differentiating the fake samples $\mathbf{l}$ from samples from the distribution of true LDR images $p_\textrm{LDR}$.
Since the samples $\mathbf{r}$ undergo stochastic projections from HDR to LDR, $G$ is forced to produce valid HDR images, resulting in the distribution $p_\textrm{HDR}^G$ of generated HDR images approaching the true distribution $p_\textrm{HDR}$ \cite{bora2018ambientgan}.

As our generator and discriminator backbone, we choose the current state-of-the-art model StyleGAN-XL \cite{sauer2022stylegan}. This model has been demonstrated to yield excellent image quality on diverse datasets. The generator model consists of two stages (left block in Fig.~\ref{Fig.structure}): First, a mapping network $M_{\theta_M}$ in the form of an MLP with parameters $\theta_M$ turns the initial random vector $\mathbf{z} \in \mathds{R}^k$ into a more disentangled latent feature representation $\mathbf{w} \in W \subset \mathds{R}^k$. Second, $\mathbf{w}$ is fed into a synthesis CNN $S_{\theta_S}$ with parameters $\theta_S$ to yield the final HDR output $\mathbf{r} = S_{\theta_S}(M_{\theta_M}(\mathbf{z}))$. Notice that the particular choice of the generator and discriminator networks is orthogonal to our approach. Except for the camera model introduced in the next paragraphs, we use StyleGAN-XL without any modifications in architecture or training hyper-parameters. We verify this in {Sec.~\ref{sec.ablation}}. 

At the core of our method is the stochastic camera model $C$ (central block in Fig.~\ref{Fig.structure}) which projects the HDR image $\mathbf{r}$ onto an LDR counterpart $\mathbf{l}$. 
It is designed to model the distribution of typical processing steps in a digital camera:

\begin{equation}
    C(\mathbf{r})
    = 
    \textrm{CRF}_{\beta, \gamma}
    \left(
    \min( 2^{\frac{e}{2}} \cdot \mathbf{r}, 1)
    \right).
\label{eq:camera_model}
\end{equation}
In the first step, we multiply each pixel of $\mathbf{r}$ with a single global exposure value.
The exposure is parameterized by the random variable $e$, capturing the exposure distribution of typical cameras arising from the combined effect of aperture, shutter speed, and sensor sensitivity (ISO). 
We do not have access to the true distribution of exposure parameters $e$, as images from in-the-wild image collections frequently do not have EXIF headers that would contain this information.
Since the exposure is a combined effect of multiple factors (aperture, shutter, ISO, time of day, etc.), we choose to model $e$ using a normal distribution, i.e.,~ $e \sim \mathcal{N}(0, \sigma_e^2)$.
The exposure variance $\sigma_e^2$ is the only hyper-parameter in our system and is analyzed in Sec.~\ref{sec.ablation}.
\begin{wrapfigure}{r}{0.2\textwidth}
    \includegraphics[width=0.2\textwidth]{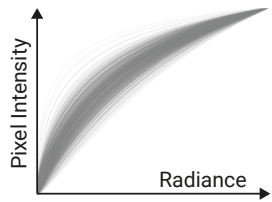}
    \vspace*{-7mm}
    \caption{CRF Distribution.}
    \vspace{-3mm}
    \label{Fig.crf} 
\end{wrapfigure}

After applying a random exposure, the $\min$ operator in Eq.~\ref{eq:camera_model} clips large radiance values to 1, effectively flattening all definition in overexposed regions. Notice that this loss of information is precisely the reason why $C$ is not invertible for individual images: By selecting an exposure, we only observe a bracket of radiance values. In contrast, our method seeks to invert $C$ over the distribution of images and camera models \cite{bora2018ambientgan}.

The final component of our model in Eq.~\ref{eq:camera_model} is the non-linear sensor response. 
The CRF describes the mapping from radiance values arriving at the sensor to pixel intensities stored in the final image.
We follow the established distribution of Eilertsen et~al. \cite{eilertsen2017hdr}:

\begin{equation*}
    \mathrm{CRF}_{\beta, \gamma}(x) = \frac{(1+\beta) x^{\gamma}}{\beta + x^{\gamma}},
\end{equation*}
with ${\beta \sim \mathcal{N}(0.6, 0.1)}$ and ${\gamma \sim \mathcal{N}(0.9, 0.1)}$ as obtained from the analysis of a large dataset \cite{grossberg2003space}. We visualize the distribution of CRFs arising from this model in Fig.~\ref{Fig.crf}.


With the full stochastic camera model in place, our system can be trained in an adversarial fashion from scratch without further modifications to the standard GAN pipeline to yield an HDR generator trained only on an LDR image dataset.

\begin{figure*}[t!]
	\centering
 \includegraphics[width=\textwidth]{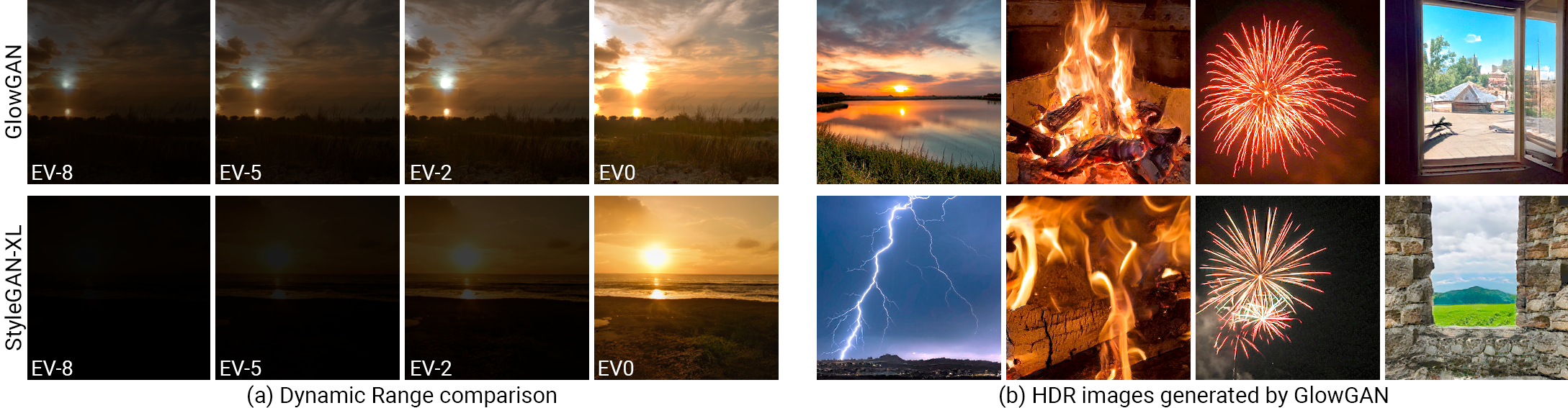}
 \vspace{-0.6cm}
	\caption{(a) Dynamic range comparison between our \HDRGAN{} (top) and a vanilla GAN (bottom) when varying the exposure values (EV) of the image. (b) HDR samples generated by our \HDRGAN{} for the five tested datasets (please see Fig.~\ref{fig:teaser} for more samples). Our method generates high quality images with much higher dynamic range and without overexposure.
    } 
     \vspace{-0.3cm}
	\label{Fig:GeneratedSamples} 
\end{figure*}

\subsection{Unsupervised Inverse Tone Mapping}
\label{sec-itm}

In addition to producing HDR image samples, a trained \HDRGAN{} can be used to perform unsupervised inverse tone mapping (ITM). While state-of-the-art ITM approaches typically rely on supervision from LDR--HDR image pairs, our method allows for recovering HDR images from their LDR counterparts without HDR data using GAN inversion \cite{zhu2016generative, xia2022gan}.

To obtain high-quality ITM results and to facilitate multi-modality, we choose a per-image optimization-based approach to perform the inversion \cite{creswell2018inverting, abdal2019image2stylegan}.
Given an LDR image $\mathbf{\hat{l}}$, we consider the following optimization objective:
\begin{equation}
    \left[
    e^*, \mathbf{w}^*, \theta_S^*
    \right]
    =
    \argmin_{e, \mathbf{w}, \theta_S}
    \Phi
    \left(
        C(S_{\theta_S}(\mathbf{w})),
        \mathbf{\hat{l}}
    \right).
\label{eq:itm-opt}
\end{equation}

Here, we jointly optimize over exposure $e$ and the latent code $\mathbf{w}$ while fine-tuning the synthesis network parameters $\theta_S$ to obtain a faithful match between the target LDR image $\mathbf{\hat{l}}$ and its reconstruction using our pipeline. $\Phi$ denotes the discrepancy measure between the two images.
Using the Adam optimizer \cite{kingma2014adam} with standard parameters, we proceed in two stages:
In the first stage, we exclude the generator weights $\theta_S$ from the optimization and measure image discrepancy $\Phi$ using the LPIPS perceptual distance \cite{zhang2018perceptual}.
In the second stage, we only optimize (fine-tune) $\theta_S$ using the pivotal tuning technique of Roich et~al. \cite{roich2021pivotal}, with $\Phi$ being the sum of a pixel-wise $\ell_2$ loss and the LPIPS perceptual distance.
Following most previous work, we relax $\mathbf{w}$ to explore the extended latent space $W^+$ \cite{abdal2019image2stylegan}.
We did not find it necessary to optimize over the CRF parameters $\beta, \gamma$ for high-quality results and consequently fix them to their mean values. More details on the optimization can be found in the \supp{supplemental}.
Upon completion of the optimization, we obtain the HDR version of $\mathbf{\hat{l}}$ via
$
    \mathbf{r^*} = S_{\theta_S^*}(\mathbf{w}^*).
$

Following the lossy projection in Eq.~\ref{eq:camera_model}, the mapping from LDR to HDR images is not unique: Overexposed regions in the LDR image can be explained by many different HDR images.
Our system allows capturing this multi-modality by running the optimization of Eq.~\ref{eq:itm-opt} multiple times with different parameter initializations for $\mathbf{w}$ and $e$.
Specifically, we initialize each optimization run with $\mathbf{w} = M_{\theta_M}(\mathbf{z})$, where $\mathbf{z}$ is a normally distributed random vector.
This allows us to obtain multiple plausible HDR solutions, which almost exclusively differ in the overexposed regions.

Obtaining pixel-accurate GAN inversion results is challenging \cite{blau2018perception, abdal2019image2stylegan}. Fortunately, in most cases, we are only interested in hallucinating content in the saturated image regions, while well-exposed pixels can be re-used after linearization. Therefore, optionally, we diminish potential distortions arising from the inversion by blending the linearized original LDR image $\mathbf{\hat{l}}$ with the reconstructed HDR result $\mathbf{r^*}$ ~\cite{eilertsen2017hdr, santos2020single}  as follows:
\begin{equation*}
    \mathbf{r^*_\mathrm{blend}}
    =
    e^* \cdot (\mathbf{m} \odot \mathbf{r^*}) + (1 - \mathbf{m}) \odot \textrm{CRF}^{-1} (\mathbf{\hat{l}} ).
\end{equation*}
Here, $\odot$ denotes the Hadamard product and $\mathbf{m}$ is a soft mask, indicating saturated pixels in $\mathbf{\hat{l}}$, which we compute for each pixel $i$ following Eilertsen et~al. \cite{eilertsen2017hdr}:
\begin{equation*}
    \mathbf{m}_i
    =
    \frac
    {\max \left( 0, \max_c \mathbf{\hat{l}}_{i,c} - \tau \right)}
    {1 - \tau},
\end{equation*}
where $\mathbf{\hat{l}}_{i,c}$ denotes the LDR image with pixel index $i$ and color channel $c$. We set the threshold $\tau = 0.97$ in all our experiments, resulting in a short ramp towards saturation.

\section{Experiments}
We have conducted the following experiments to demonstrate the effectiveness of our approach. Sec.~\ref{sec.samples} shows the generated HDR data and compares it with an LDR equivalent; 
Sec.~\ref{sec.ablation} explores the influences of the exposure distribution, model backbone, and sampled camera response curve; and
Sec.~\ref{sec.itm} presents the results of our unsupervised ITM.
We use the tone mapper of Mantiuk et~al.~\cite{mantiuk2008display} to display HDR content in this paper.
We refer readers to the \supp{supplemental} for more results and full HDR data.

\noindent\textbf{Implementation details. }
We implement our method on top of the official StyleGAN-XL~\cite{sauer2022stylegan} implementation under the PyTorch~\cite{paszke2019pytorch} environment.
Training takes six days with four RTX 8000 GPUs for 256$\times$256 resolution dataset -- roughly the same time as training a vanilla StyleGAN-XL model. 


\noindent\textbf{Datasets. }
We collect five different datasets which naturally contain scenes with high dynamic range from the Internet: Landscapes ($\sim$7700 images), Lightning ($\sim$7000 images), Windows ($\sim$4200 images), Fireplaces ($\sim$2600 images), and Fireworks ($\sim$5600 images). We randomly crop and resize each image to the target resolution. Please refer to Fig.~\ref{fig:teaser} and Fig.~\ref{Fig:GeneratedSamples} for examples of generated images from models trained on these datasets.
We collect our datasets from several websites: Flickr, Pexels, Instagram, and 500PX.
We will make the datasets available upon request.
Further, we will make all source code and pre-trained models publicly available upon publication.

\subsection{Generation of HDR Images}
\label{sec.samples}

\begin{figure}
    \includegraphics[width=\linewidth]{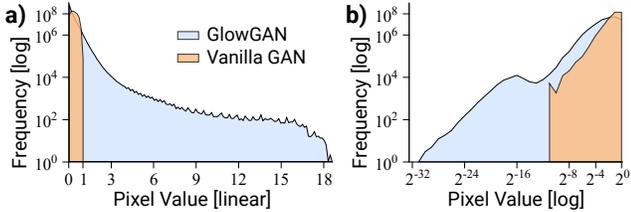}
    \caption{Pixel values from our model and a vanilla GAN. The linear-scale histogram (\emph{a}) shows that we significantly extend the dynamic range for high pixel values, while the log-scale histogram (\emph{b}, truncated to only show values up to 1) demonstrates an extension for low pixel values as well.}
    \label{Fig:histogram}  
\end{figure}

\begin{figure}[t!] 
	\centering
	\includegraphics[width=\linewidth]{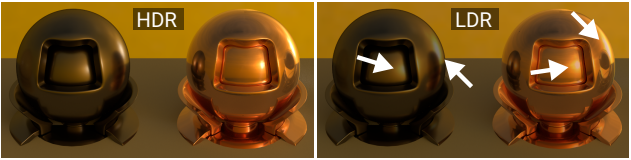}
 \caption{Image-based lighting using a generated HDR environment map (left) vs. a plain LDR equivalent (right). Arrows mark regions where the differences are most pronounced. The HDR illumination leads to high-contrast and high-frequency reflections in the specular materials (right object) and produces a more realistic bloom effect in the glossy material (left object).}
 \vspace{-0.4cm}
	\label{Fig:ibl} 
\end{figure}

We show in Fig.~\ref{Fig:GeneratedSamples} a variety of samples generated with \HDRGAN{} and a comparison with a vanilla StyleGAN-XL.
Generated samples from the vanilla GAN often bear overexposed regions similar to those in the LDR training images.
In contrast, samples from \HDRGAN{} preserve detailed appearance information even for bright objects such as the sun, as they have more extensive dynamic ranges than those from the vanilla GAN.
To show the difference in dynamic range, we plot the image brightness histogram of the two models in Fig.~\ref{Fig:histogram}, where each histogram is computed from 500 randomly sampled images.
It can be seen that the histogram of the vanilla GAN is cut off at 1, while \HDRGAN{} clearly avoids pixel intensity clamping (Fig.~\ref{Fig:histogram}a) and has a much wider histogram, also in the dark regions (Fig.~\ref{Fig:histogram}b).
These results satisfy our expectation of learning HDR information from LDR data. 
%
Moreover, we see that \HDRGAN{} can synthesize diverse images that do not exist in the real world.
It thus opens up a new avenue for getting cheap abundant HDR data which can be used in several applications, e.g., for creating environment maps for image-based lighting (IBL)~\cite{debevec2002image}, as showcased in Fig.~\ref{Fig:ibl}. Additionally, \HDRGAN{} can interpolate between two environment maps, achieving a smooth transition effect, as demonstrated in the \supp{supplemental}.



\subsection{Ablation Study}

\label{sec.ablation}
\noindent\textbf{Exposure Distribution. } 
Our method assumes the exposure $e$ follows a Gaussian distribution with variance $\sigma_e^2$. 
Here we study how $\sigma_e^2$ impacts the generated image quality and the dynamic range. 
We employ the commonly used FID~\cite{heusel2017gans} and KID~\cite{binkowski2018demystifying} scores to evaluate image quality on the Landscapes dataset.
As we do not have ground truth HDR images, the scores are computed between the generated LDR images (i.e.,, output of the camera model) and the LDR training images.
We also compute the dynamic range (DR) for each generated HDR image $\mathbf{r}$ as 
$
\textrm{DR}
=
\log_2
\left(
\sfrac
{\mathbf{r}_\textrm{max}}
{\mathbf{r}_\textrm{min}}
\right),
$
where $\mathbf{r}_\textrm{max}$ and $\mathbf{r}_\textrm{min}$ are the max and min values of the image, respectively.
We report the median and the $90^\textrm{th}$ percentile of DR computed over 50k images, referred to as DR50 and DR90, respectively.
From Table~\ref{table:dr_exploration}, we can observe a trade-off between image quality and the dynamic range, i.e., increasing $\sigma_e^2$ leads to a higher dynamic range (with diminishing returns for high $\sigma_e^2$) but slightly worse FID and KID scores.
To understand the positive correlation between $\sigma_e^2$ and DR, suppose that $\mathbf{r}$ has a low dynamic range, then with a very small or large exposure $e$ (which is more likely to happen for large $\sigma_e^2$), it would produce an out-of-distribution LDR image $\mathbf{l}$ that is overly dark or bright.
In other words, only a valid high dynamic range $\mathbf{r}$ can yield realistic LDR images when processed with different exposures.
On the other hand, as $\sigma_e^2$ increases, the camera model interferes more with the image generation process, which may increase the training difficulty as the Gaussian distribution is only an approximation to the underlying exposure distribution.
In practice, users can choose a suitable $\sigma_e^2$ depending on their goal.
In most of our experiments, we use $\sigma_e^2 = 1$ as it already removes overexposure while featuring good quality and it also obtains better scores in the inverse tone mapping application.
We provide more results on the effect of $\sigma_e^2$ in the \supp{supplemental}.
From Table~\ref{table:dr_exploration} we can further see that a vanilla StyleGAN-XL exhibits slightly higher image quality, but a substantially smaller (log-scale) dynamic range.

\begin{table}[t!]
    \caption{
    Effects of model and $\sigma_e^2$ on quality and dynamic range.
    }
    \vspace{-0.2cm}
    \centering
    \scalebox{0.9}{
    \begin{threeparttable}
	
    \begin{tabular}{lcrrrr}
         
        Model & $\sigma_e^2$ & FID$\downarrow$ & KID{\footnotesize ($\times 10^{4}$)}$\downarrow$ & DR50 & DR90 \\

        \toprule

         SG-XL\tnote{1} & -- & 3.48 & 2.69 & 8.0 & 8.0 \\
         
         Ours & 1.0  & 3.61 & 3.07 & 15.4 & 20.2 \\
         
         Ours & 3.0 & 3.87 & 4.04 & 16.2  & 20.7 \\
                  
         Ours & 5.0 & 4.00 & 4.78 & 16.5  & 20.8 \\

        \midrule

        SG2\tnote{2} & -- & 9.2 & 23.37 & 8.0 & 8.0 \\


        Ours w/ SG2\tnote{3} & 1.0 & 10.02 & 28.41 & 16.3 & 22.9 \\

        \bottomrule
     
    \end{tabular}

\begin{tablenotes}\footnotesize
\item[1] Refers to a vanilla StyleGAN-XL model.
\item[2] Refers to a vanilla StyleGAN2-ADA model.
\item[3] Refers to our approach with a StyleGAN2-ADA backbone.
\end{tablenotes}
\end{threeparttable}
}
\label{table:dr_exploration}
\end{table}



\begin{table}[t!]
	\caption{
     Comparing quality with fixed and stochastic CRFs.}
     \vspace{-0.2cm}
    \centering
    \scalebox{0.9}{
    \begin{tabular}{llrr}
          Dataset & CRF & FID$\downarrow$ &  KID{\footnotesize ($\times 10^{4}$)}$\downarrow$ \\
          \toprule
          \multirow{2}{*}{Landscapes} & Fixed & 3.89   & 3.80 \\
          & Stochastic & \textbf{3.61} & \textbf{3.07} \\
          \multirow{2}{*}{Lightning} & Fixed & 3.40  & 4.77 \\
          & Stochastic & \textbf{3.29} & \textbf{4.57} \\
          \bottomrule
    \end{tabular}
    }
    \vspace{-0.2cm}
    \label{table:crf}
\end{table}

\noindent\textbf{Generator backbone.}
We further test our method based on StyleGAN2-ADA~\cite{karras2020training}.
As Table~\ref{table:dr_exploration} shows, our method also successfully synthesizes images with high dynamic range, albeit the final image quality directly depends on the baseline generative model.

\begin{figure*}[htbp] 
	\centering
 \includegraphics[width=\textwidth]{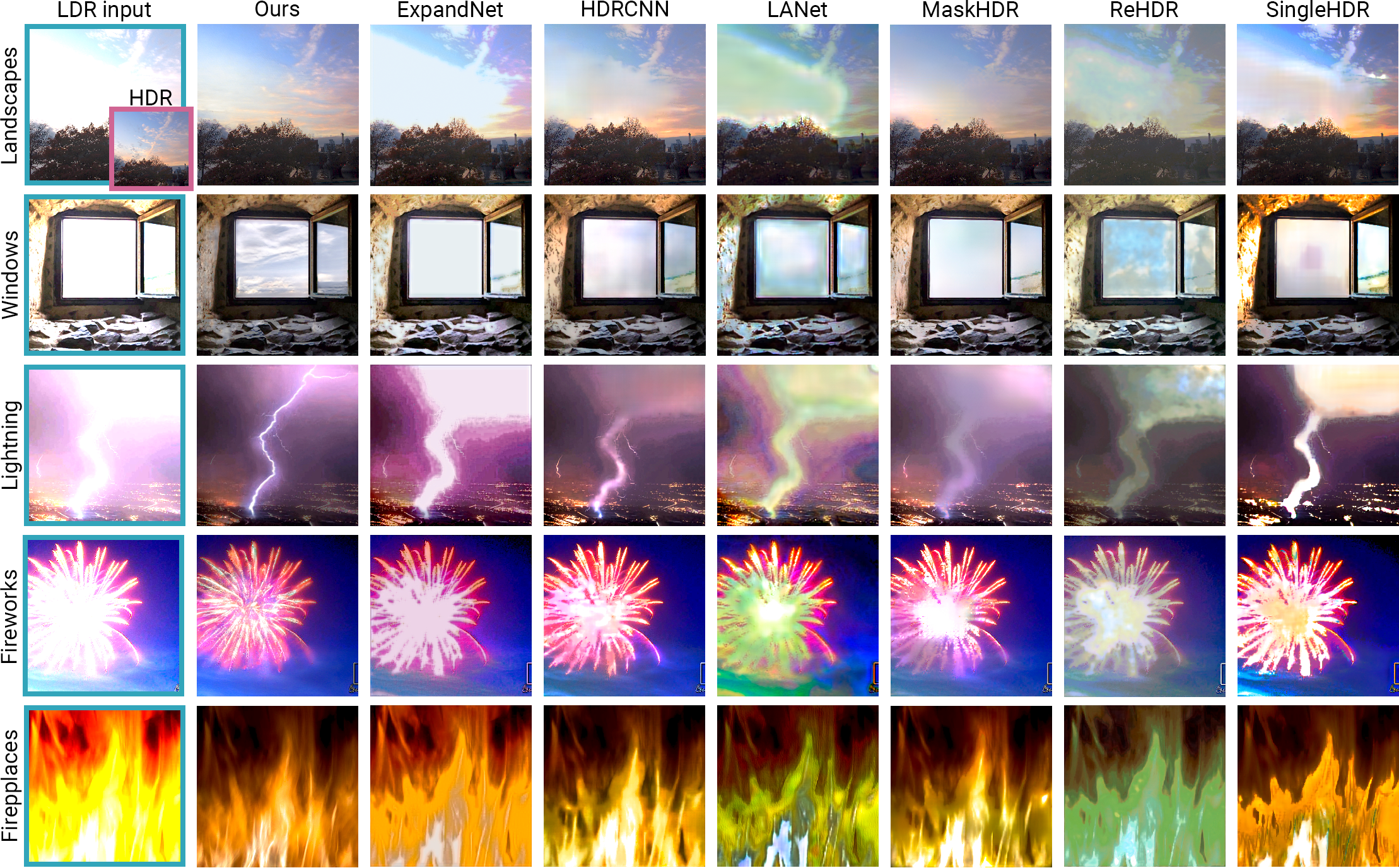}
	\caption{Results for the Inverse Tone Mapping application and comparisons to six state-of-the-art methods for different datasets. Given a single LDR image as input, our method can produce an HDR image that contains plausible and realistic content in the previously overexposed regions, while previous methods tend to produce blurred results or even noticeable artifacts in such regions. In the first row (Landscapes dataset), it can be seen that recovering the ground truth content present in the original HDR scene is not possible in fully saturated regions, however, our method is able to produce plausible results that are consistent with the scene.}
    \vspace{-0.1cm}
	\label{Fig:ITMcomparisons} 
\end{figure*}

\begin{figure}[htbp] 
	\centering
	\includegraphics[width=\linewidth]{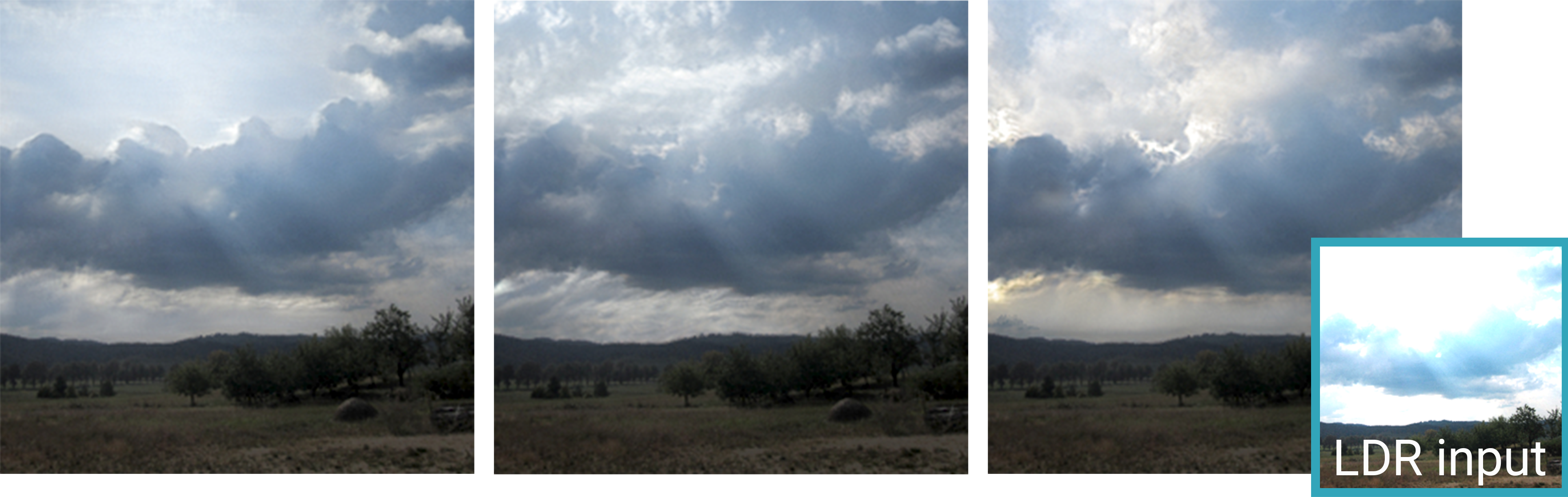}
	\caption{Our method can generate different but plausible HDR images from a single LDR input with large overexposed regions.} 
	\label{Fig:diversity} 
\end{figure}

\noindent\textbf{Stochastic CRF.}
We further study the effects of the stochastic CRF sampling process in our model.
Table~\ref{table:crf} compares our stochastic CRF sampling with a fixed CRF, using $\beta=0.6$ and $\gamma=0.9$.
Modelling the stochasticity leads to a clear improvement in FID and KID scores.
This is because the in-the-wild LDR images used for training are captured with different cameras with diverse CRFs, which can be better modelled via stochastic CRF sampling.

\begin{table*}[h!]
	\centering
    \caption{
    Evaluation of the Inverse Tone Mapping application. We achieve on-par quality with the best-performing supervised methods in reference metrics while outperforming all previous approaches in the non-reference metric that evaluates the overall naturalness and quality of the image. Note that reference metrics are not best suited for this evaluation, since the strength of our method lies in the reconstruction of missing overexposed regions, and therefore it is expected that the hallucinated content does not match that of the original HDR scene. 
    }
\vspace{-0.3cm}
\scalebox{0.9}{
    \begin{tabular}{lcrrrr}
          \multirow{2}{*}{Method} & \multirow{2}{*}{Unsupervised} & \multicolumn{3}{c}{Reference} & Non-Reference \\

          & & HDR-VDP3 $\uparrow$ & PU21-VSI $\uparrow$ & PU21-PSNR $\uparrow$ & PU21-PIQE $\downarrow$\\
        \toprule
         
         HDRCNN ~\cite{eilertsen2017hdr} & \xmark & \cellcolor{gray!16} 7.42 $\pm$ 1.02 & \cellcolor{gray!32} 0.961 $\pm$ 0.034 & \cellcolor{gray!32} 32.1 $\pm$ 5.1 & \cellcolor{gray!16} 36.1 $\pm$ 5.3 \\ 
         
         MaskHDR ~\cite{santos2020single} & \xmark & \cellcolor{gray!48} 7.60 $\pm$ 0.93 & \cellcolor{gray!48} 0.962 $\pm$ 0.032 & \cellcolor{gray!48} 32.4 $\pm$ 5.1 & \cellcolor{gray!32} 33.3 $\pm$ 6.5 \\ 
         
         SingleHDR ~\cite{liu2020single} & \xmark &7.01 $\pm$ 1.17 &0.956 $\pm$ 0.031 &30.1 $\pm$ 4.5 & 40.3 $\pm$ 6.2 \\ 
         
         ExpandNet ~\cite{marnerides2018expandnet} & \xmark &6.66 $\pm$ 1.61 &0.957 $\pm$ 0.033 &30.7 $\pm$ 4.2 & 43.2 $\pm$ 6.6 \\  
         
         ReHDR ~\cite{lee2018re} & \xmark &7.06 $\pm$ 1.31 & 0.953 $\pm$ 0.035 &30.3 $\pm$ 4.2 & 39.7 $\pm$ 4.7 \\ 
        
         LANet ~\cite{yu2021luminance} & \xmark &6.94 $\pm$ 0.98 & 0.956 $\pm$ 0.031 &29.0 $\pm$ 3.6 & 40.6 $\pm$ 6.5 \\ 

         Ours & \cmark & \cellcolor{gray!32} 7.44 $\pm$ 0.94 & \cellcolor{gray!32} 0.961 $\pm$ 0.032 & \cellcolor{gray!16} 31.8 $\pm$ 4.4 & \cellcolor{gray!48} 31.8 $\pm$ 5.1 \\ 
         \bottomrule
    \end{tabular}
    }
\label{table:itm_evaluation_withgt}
\end{table*}

\subsection{Application: Inverse Tone Mapping}
\label{sec.itm}

A potential application of our approach is unsupervised inverse tone mapping (ITM). We show both quantitatively and through a user study that our method outperforms previous approaches in hallucinating content in large overexposed regions, effectively recovering HDR content from a single LDR image. 

\paragraph{Objective comparisons.} We compare our unsupervised approach to six state-of-the-art fully-supervised ITM methods, which we abbreviate for simplicity as HDRCNN~\cite{eilertsen2017hdr}, MaskHDR~\cite{santos2020single}, SingleHDR~\cite{liu2020single}, ExpandNet~\cite{marnerides2018expandnet}, ReHDR~\cite{lee2018deep}, and LANet~\cite{yu2021luminance}. Following the work of Hanji et al. on quality assessment of single image HDR reconstruction methods~\cite{hanji2022comparison}, we select their three recommended full-reference metrics (PU21-PSNR, PU21-VSI, and HDR-VDP3) as well as their recommended non-reference metric (PU21-PIQE). As test set for the reference metrics we use as HDR ground truth a set of 62 images collected from existing datasets~\cite{hanji2022comparison, panetta2021tmo} and generate the corresponding LDR input images following the pipeline proposed by Eilertsen et al.~\cite{eilertsen2017hdr}. For the non-reference metric, we use an extended set of 100 LDR images obtained from the Internet which we use directly as input with 15$\%$ to 45$\%$ of the pixels saturated. For fairness in these comparisons, both sets are composed of landscape images, since fully-supervised approaches are typically trained with datasets mainly containing this type of content.  
We show in Table~\ref{table:itm_evaluation_withgt} the results of these metrics and in Fig.~\ref{Fig:ITMcomparisons} visual comparisons for our five datasets. Note that, since our method focuses on hallucinating content in completely saturated regions, it is highly unlikely that this content fully matches that of the original ground truth image, therefore full-reference metrics are not well suited for assessing the quality of our reconstructions. Nevertheless, our unsupervised method is still on par with previous fully-supervised approaches for the reference metrics, while it excels in the non-reference metric, showing that our hallucinated content is more plausible in terms of naturalness. Additionally, previous methods can generate only one potential reconstruction given an input LDR image, while our approach allows the generation of multiple results with different but plausible semantic information, as Fig.~\ref{Fig:diversity} shows.

\paragraph{User Study.}
\label{subsec.user_study}

\begin{figure}[t!] 
	\centering
	\includegraphics[width=\linewidth]{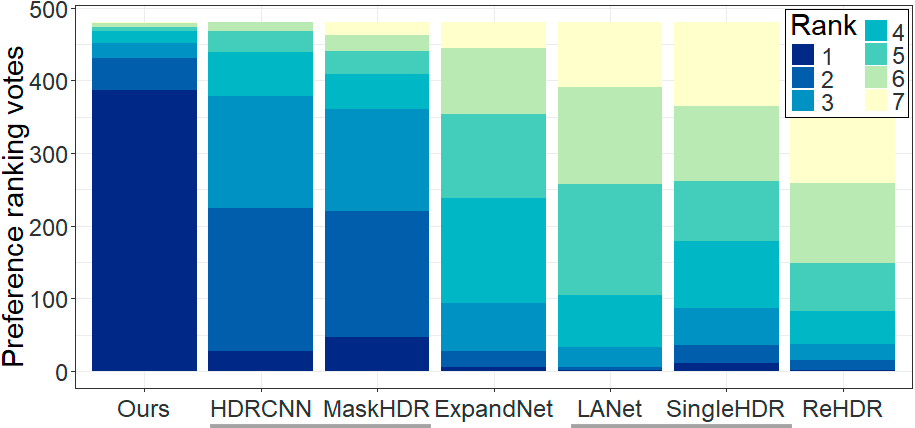}
    \vspace{-0.6cm}
	\caption{Preference rankings for the seven inverse tone mapping methods aggregated across participants and scenes. Different colors indicate the rankings (from 1 to 7). Methods marked in the same set (gray underline) are statistically indistinguishable, while all others present statistically significant differences in their distributions.} 
	\label{Fig:user_study} 
\end{figure}

Since one of the main strengths of our work is the capability to hallucinate plausible content in overexposed regions, reference metrics are unsuitable for comparisons. Additionally, the image diversity of available ground-truth HDR datasets is limited. Therefore, we perform a subjective study in order to further assess the quality of our generated results for the inverse tone mapping application. We include 20 scenes (four for each of our five datasets). For each scene, HDR results obtained with each of the seven methods were shown on a single screen, and participants were asked to rank the seven images from 1 (most preferred) to 7 (least preferred). The presentation order of the scenes and methods was randomized. The images were displayed in an HDR display Dell UP3221Q (3840$\times$2160 resolution) in a standard office room with natural illumination, and participants sat at a distance of 0.5 meters from the display. A total of 24 participants (38$\%$ female, aged 22 to 37 years old with normal or corrected-to-normal vision) participated in the study. We show in Fig.~\ref{Fig:user_study} the preference rankings for each method, aggregated for all scenes and participants. To analyze the results, we use pairwise Kruskal-Wallis tests adjusted by Bonferroni correction for multiple comparisons since the rankings do not follow a normal distribution. Results reveal that our method was ranked significantly higher than all others ($p < 0.001$), and it was selected as the top performing method in over 80$\%$ of the trials. 

\section{Conclusion \& Discussion}

We have introduced \HDRGAN{}, a novel paradigm for learning HDR imagery from LDR data. Our method is orthogonal to other advances in generative adversarial learning and can be easily incorporated into any GAN-based pipeline. A trained \HDRGAN{} acts as a strong prior, producing starkly more plausible inverse tone mapping results than previous approaches. 
\begin{wrapfigure}{r}{0.2\textwidth}
    \includegraphics[width=0.2\textwidth]{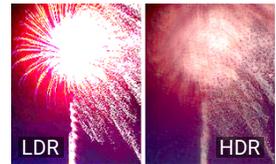}
    \vspace*{-7mm}
    \caption{Failure case.}
    \vspace*{-3mm}
    \label{Fig.failure} 
\end{wrapfigure}
Our inverse tone mapping method builds on GAN inversion via optimization, which can sometimes result in low-quality images, especially for high-frequency content (Fig.~\ref{Fig.failure}) -- a problem that is orthogonal to our approach.
We heavily rely on training data with a diverse exposure distribution. While this assumption is oftentimes naturally satisfied for in-the-wild photo datasets, the dynamic range we can obtain is tightly linked to the exposure variance in the dataset. We hope that our approach inspires future work on learning rich models from casually captured images.

{\small
\bibliographystyle{ieee_fullname}
\bibliography{egbib}
}

\end{document}